\documentclass[sigconf]{acmart}
\usepackage{multirow}
\AtBeginDocument{%
  \providecommand\BibTeX{{%
    \normalfont B\kern-0.5em{\scshape i\kern-0.25em b}\kern-0.8em\TeX}}}

\copyrightyear{2020}
\acmYear{2020}
\setcopyright{rightsretained}
\acmConference[ICAIF '20]{ACM International Conference on AI in Finance}{October 15--16, 2020}{New York, NY, USA}
\acmBooktitle{ACM International Conference on AI in Finance (ICAIF '20), October 15--16, 2020, New York, NY, USA}
\acmDOI{10.1145/3383455.3422566}
\acmISBN{978-1-4503-7584-9/20/10}

\begin{document}

\title{Graphical Models for Financial Time Series and Portfolio Selection}

\author{Ni Zhan}
\authornote{Both authors contributed equally to this research.}
\email{nzhan@andrew.cmu.edu}
\author{Yijia Sun}
\authornotemark[1]
\email{yijias@andrew.cmu.edu}
\author{Aman Jakhar}
\author{He Liu}
\affiliation{%
  \institution{Carnegie Mellon University}
}







\renewcommand{\shortauthors}{Zhan and Sun, et al.}

\begin{abstract}
 We examine a variety of graphical models to construct optimal portfolios. Graphical models such as PCA-KMeans, autoencoders, dynamic clustering, and structural learning can capture the time varying patterns in the covariance matrix and allow the creation of an optimal and robust portfolio. We compared the resulting portfolios from the different models with baseline methods. In many cases our graphical strategies generated steadily increasing returns with low risk and outgrew the S\&P 500 index. This work suggests that graphical models can effectively learn the temporal dependencies in time series data and are proved useful in asset management.
\end{abstract}





\maketitle

\section{Introduction}
Portfolio selection is a common problem in finance. In general, investors wish to maximize returns while  minimizing risk. Markowitz theory suggests that portfolio diversification minimizes risk and the optimal portfolio takes into account correlated movements across assets. Existing works on porfolio selection use the historical covariance matrix of returns. However, factors such as market index, sector, industry and other stocks and commodities that may affect asset cash flows can result in a high degree of correlation among equities. This causes the historical covariance matrix to be ill conditioned and the optimal portfolio highly sensitive to small changes. Expanding the universe of assets also requires a greater amount of data to estimate the covariance matrix. Furthermore regime changes and the non-stationary nature of financial markets discourage the use of static covariance matrices. 

From a graph viewpoint, estimating the covariance using historic returns models a fully connected graph between all assets.
The fully connected graph appears to be a poor model in reality, and substantially adds to the computational burden and instability of the problem.

The goal of this work is to develop graphical models that can capture the time varying patterns in the covariance matrix and reflect the cross-series dynamics at multiple time indices.
Using graph inference algorithms and thresholding, we plan to discover and incorporate the factor dependencies in a partially connected graph. Therefore, we leverage graphical models that are able to reflect temporal changes among stocks thus addressing the issue of correlations changing over time. Within a time period, a graph allows selection of independent assets for the portfolio, which should improve robustness of our solution.  

In order to learn the overall time series features, we use principle component analysis (PCA) and autoencoders to capture the latent space distribution. We employ variational autoencoders with Gaussian and Cauchy priors to model temporal dependencies and reflect multi-scale dynamics in the latent space.
Additionally, we construct a sequence of graphical models using dynamic clustering techniques and structural learning. We associate a graphical model to each time interval and update the graph when moving to the next time point. We use price data of US equities from the S$\&$P 500 index to construct graphical models to create portfolios, simulate returns, and compare with benchmarks. 

\section*{Related Work}
Various graph methods have been used for the portfolio selection problem. The literature includes many examples of 
variance-covariance networks that analyze complicated interactions and market structure of financial assets~\cite{varcovPort,meanvarport,baysianPort}. Liu et. al. used elliptical-copula graphical models on stock returns, and used the graphs to select independent stocks \cite{liu-2015-semip-graph}. 
The paper chose elliptical-copula graph model over Gaussian graphical models because elliptical-copula models better model heavy tail distributions common in finance. 
The paper shows that graphs modeling asset “independence” can be learned from historical returns and used for effective portfolio strategies, however it did not consider dynamic graphs.

\begin{figure}[h]
    \centering
    \includegraphics[width=0.25\textwidth]{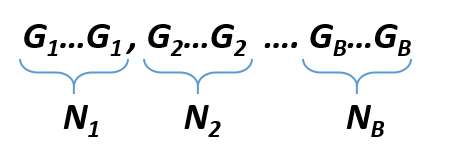}
    \caption{Blocks of Gaussian graph sequences}
    \label{fig:blocks-of-graphs}
    \Description{Blocks of Gaussian graph sequences.}
\end{figure}

Previously, time varying behavior was modeled as dynamic networks whose topology changes with time. Talih and Hengartner proposed a graphical model for sequences of Gaussian random vectors when changes in the underlying graph occur at random times, mimicking the time varying relationships among collection of portfolios ~\cite{talih-2005-struc-learn}. 
Time series data is separated into a pre-determined number of blocks. The sample precision matrix estimated within each block serves as the foundation to construct the time-varying sequences of graphs which arises in blocks as shown in Fig.~\ref{fig:blocks-of-graphs}.
The graphs $G_b$ and $G_{b+1}$ in successive blocks differ only by the addition or deletion of at most one edge.
Markov Chain Monte Carlo is used to recover the segmented time varying Gaussian graphical sequences. Experiments using this model were done to estimate portfolio return in five U.S. industries. However, this paper assumed that the total number of distinct networks are known a priori and the network evolution is restricted to changing at most a single edge at a time. Additionally, dimensionality increases when temporal resolution is small, imposing significant computational burden.

As an extension to dynamic dependence networks, Isogai~\cite{isogai2017} proposed a novel approach to analyze a dynamic correlation network of highly volatile financial asset returns by using network clustering algorithms to mitigate high dimensionality. Two types of network clustering algorithms~\cite{isogai2014,isogai2016} were employed to transform correlation network of individual portfolio returns into a correlation network of group based returns. Groups of correlation networks were further clustered into only three representative networks by clustering along the time axis to summarize information on the intertemporal differences in the correlation structure. A case study was conducted on Japanese stock dataset. Other studies used autoencoders to dimensionally reduce stock returns \cite{gu-2019-autoen-asset,heaton-2016-deep-learn-finan}. 

\section*{Methods}
The primary objective of this work is to exploit a collection of graphical models to analyze dynamic dependencies among stocks and aid trading strategies. We would like to have a better understanding of the stock features by learning the latent space of stock time series. With a good latent space representation, stocks with similar features will fall into the same cluster. To capture the time-varying correlation among stocks, we divide the overall time horizon into multiple time intervals, zoom in to each interval and construct a local graphical model, and update the local connectivities as we move on to the next temporal period. In each time interval, we aim to identify a suitable number of stock clusters based on the graphical model and develop a portfolio selection strategy. After selecting our portfolios, we test their performance using a backtest simulation and market data. 

We employ a few different clustering and structural learning techniques, including PCA, autoencoders, agglomerative clustering, affinity propagation clustering, and graphical LASSO, to create portfolios that maximize return and minimize risk. 



\subsubsection*{Dataset}


We used price data of US equities from tech, financials, and energy sectors of the S$\&$P 500 index. The data matrix has available stocks as data observations and daily returns as features, i.e. the data matrix has shape [number of stocks, number of daily returns]. Data was obtained on a closing basis at a daily frequency for January 1st, 2012 until January 1st, 2020. We excluded stocks which had missing data for this time period. We used adjusted close to account for splits, dividends etc.
The daily closing price data for the stocks and other financial variables was obtained using yahoo finance for python.

\subsubsection*{PCA}
We used empirical returns of one year prior as training data for the simulated year. We used PCA to dimensionally reduce the data matrix of the stocks to three components. We inverse transformed stock representations in reduced space to full data, and calculated the L-2 norm of difference between actual data and recovered data for each stock. The ten stocks with the largest difference in L-2 norm were selected for "Max Difference" portfolio while ten stocks with the smallest difference were selected for "Min Difference" portfolio. PCA extracts information about the stock returns, and stocks with large difference between recovered and actual data indicate unexpected or "difficult to capture" behavior.

\subsubsection*{Autoencoders}

To extend the dimension reduction method and capture more complex interactions, we tested two autoencoders. The observed variables $\mathbf{x}$ for the autoencoders are empirical returns of the selected stocks. The latent space $\mathbf{z}$ has two dimensions. The variational distribution  $q_\phi(\mathbf{z} | \mathbf{x})$ approximating $p(\mathbf{z} | \mathbf{x})$ is assumed Normal for both autoencoders. The likelihood $p_\theta(\mathbf{x}|\mathbf{z})$ is assumed Normal for one autoencoder, and Cauchy for the other. Specifically, for the Normal autoencoder, $p_\theta(\mathbf{x} | \mathbf{z}) = N(\mathbf{x}; \mu_\theta(\mathbf{z}), \Sigma^2_\theta (\mathbf{z}))$, and for the Cauchy autoencoder,  $p_\theta(\mathbf{x} | \mathbf{z}) = \text{Cauchy}(\mathbf{x}; x_{0,\theta}(\mathbf{z}), \gamma_\theta (\mathbf{z}))$. For both autoencoders, the recognition and generative networks are parameterized by neural nets with one hidden layer with 100 reLU neurons and latent space of two dimensions. We chose a Cauchy distribution because stock return distributions usually have heavy tails, and therefore expect the Cauchy autoencoder to have more reliable results. To select portfolios, we found the latent space representation of each stock, and calculated the L-2 norm of difference between real data and generated data (from latent space) for each stock. The ten stocks with max and min L-2 norms were selected for Max and Min Difference portfolios, respectively, similar to the PCA strategy.

\subsubsection*{Dynamic clustering and graphical sequences}

To better capture how stocks move in relation to one another throughout the time period, we utilize clustering and structural learning techniques to create dynamic graph structures. Even though the dataset contains stocks from three sectors, within the same sector, some stocks are more correlated than the others, and stocks from different sectors can also have non-negligible correlations. The goal here is to identify the most suitable number of clusters to assign stocks from three sectors into. Two clustering techniques are adopted here: agglomerative clustering and affinity propagation clustering. 

Agglomerative clustering divides stocks into a number of clusters according to pair-wise Euclidian distance. It requires the number of clusters to be pre-determined. We used hierarchical clustering with Ward's minimum variance criterion to produce a dendrogram which in turn is used to determine the number of clusters by drawing a horizontal line and counting the number of vertical lines it intercepts.
A total of 15 clusters were used for agglomerative clustering. 

Affinity propagation~\cite{aff_prop} is a clustering technique that does not require an input number of clusters. It relies on similarity calculation between pairs of data points to determine a subset of representative examples in the dataset. The similarity between two points satisfies that $s(x_i,x_j)>s(x_i,x_k)$ if and only if $x_i$ is more similar to $x_j$ than to $x_k$. A responsibility matrix $\mathcal{R}$ and availability matrix $\mathcal{A}$ serve as message exchanging paths between data points. Clusters are updated by alternating between the responsibility matrix update and availability matrix update.
\begin{eqnarray*}
r(i,k)&=& s(i,k)-\max_{k'\neq k} \{a(i,k')+s(i,k')\}\\
a(i,k)&=&\min \big(0,r(k,k)+\sum_{i'\notin \{i,k\}}\max(0,r(i',k))\big)\; i\neq k\\
a(k,k)&=&\sum_{i'\neq k}\max(0,r(i',k))
\end{eqnarray*}
This approach is able to identify a high quality set of exemplars and the corresponding clusters with much lower error and lower computational burdens compared to agglomerative clustering. The number of clusters is flexible and updated throughout the time horizon. It is suitable to identify clusters when the data size is large. 

\begin{figure}[h]
    \centering
    \includegraphics[width=0.47\textwidth]{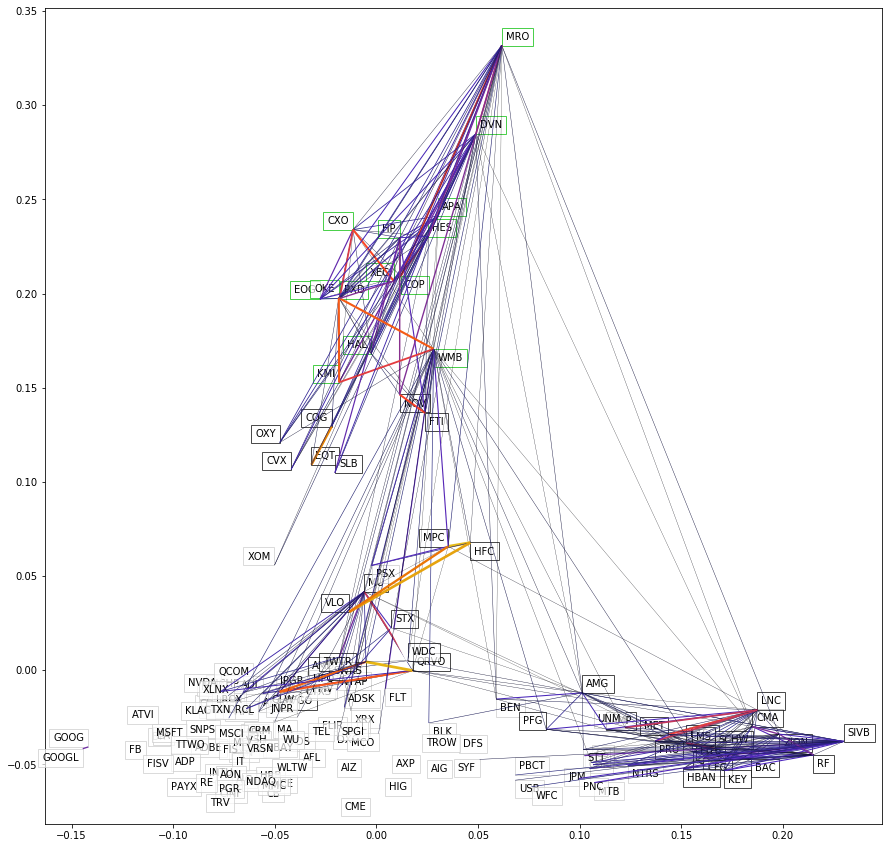}
   \caption{An example of connected graphs from dynamic clustering}
    \label{fig:graph}
    \Description{An example of connected graphs from dynamic clustering.}
\end{figure}

The data matrix contains daily returns of each stock, and has shape [number of stocks, number of daily returns]. We used spectral embedding on the daily returns to transform the stocks to a 2D plane and reduce dimensionality. An example of the 2D embedding of stocks is shown in Fig. \ref{fig:graph}. Edge connectivities were created using graphical LASSO with thicker edge indicating stronger correlation. 
At the beginning of each quarter, we rely on daily returns from the previous quarter, construct a lower-dimensional embedding space, and generate clusters using the before-mentioned clustering techniques. The edge connectivities from graphical LASSO are connectivity input for agglomerative clustering. A new clustering is created at the beginning of each quarter and therefore updated throughout the time horizon quarterly.

For agglomerative clustering and affinity propagation portfolios, the portfolio selection strategy is as follows. For each cluster, the top ten stocks with minimum Euclidean distance (in the embedding space) to cluster centers were added to the portfolio. If a cluster had fewer than ten stocks, no stocks were added to the portfolio from that cluster. Because the clusters were created quarterly, the portfolios were also selected quarterly.

To benchmark the model performance, we also constructed portfolios using PCA followed by KMeans clustering. KMeans with static clusters as well as quarterly updated clusters were both used as benchmarks. In all cases with KMeans, we used a fixed number of clusters (ten clusters), and the stock with the minimum Euclidean distance to each cluster center was part of the portfolio. Therefore KMeans portfolios had ten total stocks.

\subsection*{Testing Frameworks}

We used portfolio rebalancing strategy to test the performance of stock selection. This method rebalances the portfolio at some frequency (we used monthly). The weights across the portfolio's stocks are either equal weights or determined from mean-variance optimization (described below). The portfolio is initialized with weights of the selected stocks. At each rebalance time point, shares are bought or sold to renormalize the dollar amounts by weight across the stocks. We compared the PCA and autoencoder portfolio selection strategy with rebalancing and buying and holding the S\&P 500. Metrics to evaluate strategies included total returns, daily return standard deviations, and Sharpe ratios across simulation time-periods. Note that high Sharpe is desirable and indicates high returns with lower risk.
We followed the equal weight rebalancing strategy to test the performance of dynamic clustering, comparing affinity propagation, agglomerative clustering, PCA KMeans portfolios and the S\&P 500. At the end of each quarter, stock clusters were updated by selling the existing portfolio and buying stocks from the new cluster with equal dollar amounts. 

\subsubsection*{Efficient frontier weights} Efficient frontier weights were determined using PyPortfolioOpt, with expected means and covariance calculated from empirical returns of the year prior to simulation. The solver optimized for maximum Sharpe, and stock weights were unconstrained between 0 to 1.

\section*{Results}

\subsection*{PCA, Autoencoder Portfolios}
The two autoencoders were trained with Auto-encoding Variational Bayes (AEVB) that optimizes a stochastic estimate of evidence lower bound objective (ELBO) \cite{kingma-2013-auto-encod}. The autoencoders were trained for over 200 epochs, and the lower bound of log-likelihood converged. The training for each year was repeated ten times as the training is stochastic. The max and min difference portfolios across the ten trainings were aggregated per year, and the ten stocks which appeared most frequently in a year were used for simulating the following year.

The PCA and autoencoders portfolio selection and monthly rebalance strategy was implemented for five simulation years, 2014-2018 inclusive. We rebalanced using both equal weights and efficient frontier weights. To compare against simpler methods, we constructed additional portfolios: volatility (Vol) and "average return over volatility" (AvgRet/Vol). The max and min Vol portfolios had ten stocks with highest or lowest standard deviation of returns, respectively, in year prior to simulation. 
A stock selected for Max PCA or autoencoder portfolio has a large error in its model representation, and may have high volatility. Therefore we wanted to compare volatility alone with PCA and autoencoder portfolios. The "average return over volatility" represents a proxy for Sharpe ratio, and we wanted to test if individual stocks with high Sharpe combined would have good portfolio performance. Table \ref{table:pca-autoenc-portf2} shows the simulation results: yearly returns (\%) and standard deviation of daily returns (\%). Daily return standard deviations are higher for max portfolios than min portfolios for PCA, autoencoder, and Vol, which is expected based on their construction. Table \ref{table:pca-autoenc-portf2} also includes the average return and Sharpe ratio over the five simulation years. Risk free return in Sharpe ratio was one-year Treasury yield averaged for that year. Results for 2019 are reported as a forward-test (the data was completely withheld prior to reporting), and holding S\&P 500 is shown for comparison.

From Table \ref{table:pca-autoenc-portf2}, there are several observations which show the PCA and autoencoder strategies are useful in portfolio selection and able to create portfolios with higher return at lower risk, more so than volatility or AvgRet/Vol alone. The Max PCA and Max autoencoder portfolios perform better or on-par with Max Vol in terms of yearly return and Sharpe, from 2014-2018. For PCA and autoencoder, Max outperforms Min in both Sharpe and average returns, which is not the case with Vol. Specifically Min Vol has higher Sharpe but lower average returns than Max Vol. The Max AvgRet/Vol portfolio has higher Sharpe but lower average return than the other portfolios, and is not always able to obtain an optimal weights solution. Therefore, the model-based strategies aid in choosing portfolios which have better returns, Sharpe, and weights optimization capability, over Vol or AvgRet/Vol alone.

Other observations are that PCA seems to have lower risk than autoencoder. This is likely because PCA is a simpler model and stochasticity was introduced in autoencoder training. Improvements to the autoencoder model can be considered for future work. For some returns such as 309\% and 142\%, the efficient frontier allocated all assets into one stock when given ten stocks. Picking one outperforming stock can give extraordinarily high returns compared with selecting multiple stocks for a portfolio. In the cases with 309\% and 142\% returns, the efficient frontier optimizer selected only one stock because portfolio weight allocations were unconstrained between 0 and 1. Future work can include constrained weight allocation for the efficient frontier portfolios. In 2018, the general market trended down, and Min Vol portfolio performed the best, while years 2014 through 2017 were bull markets. The best portfolio selection strategy is likely different depending on the overall market trend. It would be interesting to consider optimal strategies for different market trends and transitions.

\begin{figure}[h]
    \centering
    \includegraphics[width=0.47\textwidth]{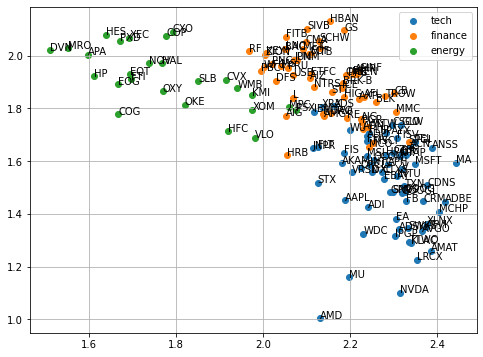}
    \caption{Latent space representation from a trained Cauchy autoencoder}
    \label{fig:c-latent-sp}
    \Description{Latent space representation from a trained Cauchy autoencoder groups sectors together.}
\end{figure}

Table \ref{table:pca-autoenc-portf2} included Normal autoencoder results only because the Cauchy autoencoder had similar  results. We examined the latent space representations of the stocks. Between PCA and the two autoencoders, the Cauchy autoencoder had the best performance in separating stocks by sector, shown in Fig \ref{fig:c-latent-sp}. This shows the model learned relevant information about the stocks. We find it quite remarkable that the autoencoder clustered the sectors considering the only training data was daily stock return for a year. It may show that stock returns are quite correlated within sectors. 

The strategies commonly selected some stocks while others were more specific to certain models. For example, for the Min portfolios, BRK-B, USB were common across Vol, PCA, and autoencoders. AFL was more often selected by Cauchy autoencoder, MMC by Vol, BLK by Normal autoencoder. For the Max portfolios, AMD, MU were common across Vol, PCA, and autoencoders, while AKAM was more specifically selected by PCA. The fact that some stocks were common across all models shows that models learned relevant information, while some stocks specifically selected by certain models shows that models learned distinct information. 
In addition, Max portfolio stocks exhibited "outlier" behavior. 

\newcommand{\STAB}[1]{\begin{tabular}{@{}c@{}}#1\end{tabular}}

\begin{table*}[!htbp]
\centering
\caption{PCA, Autoencoder, Volatility Portfolio returns and standard deviations (\%) by simulation year}
\label{table:pca-autoenc-portf2}
\begin{tabular}{lp{1.1cm}lrrrrrp{1cm}p{1cm}lp{.7cm}}\toprule                       &      &        &      2014 &       2015 &       2016 &     2017 &     2018 & Avg Yr Ret &   Daily Ret Std &      Sharpe & 2019 Test \\\midrule \multirow{8}{*}{\STAB{\rotatebox[origin=c]{90}{Equal Weights}}} & PCA & Max    &       47.7 &  10.4 & 65.4 &  34.2 & -3.1 & 30.9 & 1.39 & 1.19 & 36.5\\                       &      & Min    &     10.7 &   1.69 &  25.7 &  19.6 & -16.8 & 8.2 & 0.94 & 0.48 & 32.3\\             & AutoEnc & Max &  29.8 &  3.85 &  75.8 &  25.3 & -5.02 & 26.0 & 1.5 &  0.88 & 28.2\\                       &      & Min &  12.9 &  -1.66 &  23.5 &  16.0 & -20.7 & 6.0 &1.05 & 0.31 & 33.4\\& Vol & Max &  46.0&  -0.86 &  74.4 &  18.9 & -5.52 & 26.6 &1.57 &  0.84 & 37.4\\                       &      & Min & 5.54 &  2.64 &  21.7 &  22.8 & 3.54 &11.2 & 0.79 & 1.13 &31.2\\& \multirow{2}{1cm}{AvgRet/ Vol} & Max &  16.9 &  6.29 &  21.3 &  24.3 & 0.72 & 13.9 & 1.12 &  1.39 & 40.6\\                       &      & Min &  12.5 &  -18.5 &  33.5 &  38.9& -22.7 & 8.7 & 1.27 & 0.30 & 2.0\\\midrule  \multirow{6}{*}{\STAB{\rotatebox[origin=c]{90}{Eff Frontier}}}
 & PCA & Max    &       60.5 &  19.5 &  25.9 &  65.5 & -14.1 & 31.5 & 1.94 & 1.03 & 142\\
 &      & Min    &     22.3 &   -6.12 &  27.6 &  25.8 & -9.66 & 12.0 & 1.05 & 0.66 & 44.2\\& AutoEnc & Max &  58.4 &  19.5 &  309 &  65.5 & -13.4 & 87.8 & 2.66 &  0.76 & 141 \\                       &      & Min &  15.9 &  2.11 &  35.4 &  28.1 & -10.2 & 14.3 & 1.18 & 0.78 & 11.6\\& Vol & Max &  57.7 &  19.5 &  25.9 &  20.4 & -13.3 &22.0 & 2.0 &  0.91 & 142\\                       &      & Min &  11.6 &  -1.93 &  28.6 &  15.2 & 3.49 & 11.4 & 0.89 & 0.98 & 27.7\\\midrule\multicolumn{3}{c}{S\&P} & 14.4 &  1.29 &  13.6 &  20.7 & -5.13 & 9.0 & 0.79 & 0.82 & 30.4\\
\bottomrule\end{tabular}
\end{table*}

\subsection*{Dynamic Clustering}

\begin{figure*}[h]
    \centering
    \includegraphics[width=0.8\textwidth]{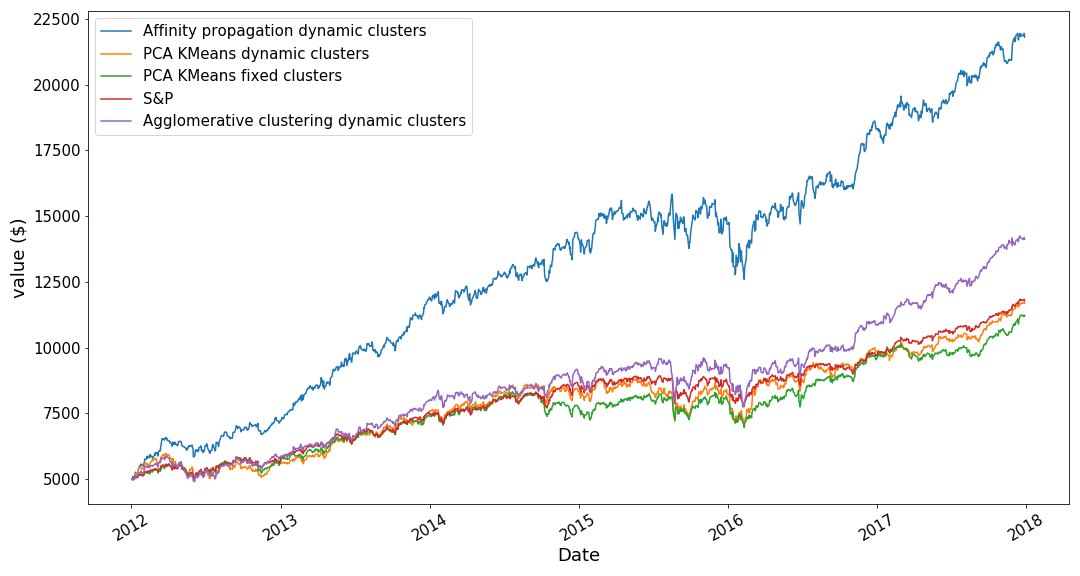}
    \caption{Rebalancing with different clustering strategies vs. S\&P 500 from 2012 to 2018 }
    \label{fig:allvsp}
    \Description{Affinity propogation dynamic clusters outperforms other strategies and benchmarks in simulation test from 2012 to 2018.}
\end{figure*}

We used trading data from January 1st, 2012 to January 1st, 2019 to evaluate the performance of the model. On the first day of each quarter, clusters are formed following agglomerative and affinity propagation clustering techniques using daily closing price from the previous quarter, and new clusters are created each quarter. A portfolio is created by selecting stocks with minimum Euclidean distance to cluster centers in the lower-dimensional transformed space. We also generated two portfolios using PCA with KMeans clustering, with fixed clusters and quarterly updated clusters respectively, as our benchmark portfolios. Figure~\ref{fig:allvsp} compares the various portfolios using monthly rebalance strategy and holding the S\&P 500. Both KMeans cluster portfolios lead to comparable performance with the S\&P 500. KMeans with dynamic cluster update performs better than using the same clusters throughout the time horizon. Both agglomerative clustering and affinity propagation outperformed S\&P 500 index, with affinity propagation generating the highest returns throughout seven years. Since affinity propagation identifies the most suitable number of clusters for the given data set, the resulting cluster size can differ from what is used in agglomerative and KMeans clustering. The likelihood of a stock being assigned to the correct cluster is higher for affinity propagation, therefore stocks that constantly beat the S\&P 500 are likely to be assigned to the same cluster and correspond to minimum distance in the spectral embedding space. Creating portfolios using such stocks steadily outgrows the performance of S\&P 500. We found that both cluster update and cluster size can influence the quality of the portfolio. Involving temporal changes to the clusters invariably boost the performance. 

\section*{Conclusion}

We showed that graphical models learn useful information and correlation between stocks only based on their returns. We developed a portfolio selection using PCA and autoencoder models and rebalance strategy that selects high return, low risk portfolios. We also explored the effect of dynamic clustering on overall portfolio returns. We observed that dynamic cluster update yields higher returns than using fixed clusters. A flexible cluster size also improves the performance than using a constant cluster size. When stocks are assigned to the correct clusters throughout the time horizon, with rebalancing strategies that minimizes risk, we are able to create a portfolio with steadily increasing returns. For future work, we can include more data into the analysis and model training, such as using trading volume, expanding number of years and stock sectors. There are other experimental factors which can be varied such as dimensions in latent space, stock selection strategy, rebalance frequency and timing. There are many questions to explore, and this work shows that graphical models have interesting and useful applications in asset management.

\bibliographystyle{ACM-Reference-Format}
\bibliography{acmart}


\begin{thebibliography}{12}


\ifx \showCODEN    \undefined \def \showCODEN     #1{\unskip}     \fi
\ifx \showDOI      \undefined \def \showDOI       #1{#1}\fi
\ifx \showISBNx    \undefined \def \showISBNx     #1{\unskip}     \fi
\ifx \showISBNxiii \undefined \def \showISBNxiii  #1{\unskip}     \fi
\ifx \showISSN     \undefined \def \showISSN      #1{\unskip}     \fi
\ifx \showLCCN     \undefined \def \showLCCN      #1{\unskip}     \fi
\ifx \shownote     \undefined \def \shownote      #1{#1}          \fi
\ifx \showarticletitle \undefined \def \showarticletitle #1{#1}   \fi
\ifx \showURL      \undefined \def \showURL       {\relax}        \fi
\providecommand\bibfield[2]{#2}
\providecommand\bibinfo[2]{#2}
\providecommand\natexlab[1]{#1}
\providecommand\showeprint[2][]{arXiv:#2}

\bibitem[\protect\citeauthoryear{Buser}{Buser}{1977}]%
        {meanvarport}
\bibfield{author}{\bibinfo{person}{Stephen~A. Buser}.}
  \bibinfo{year}{1977}\natexlab{}.
\newblock \showarticletitle{Mean-Variance Portfolio Selection with Either a
  Singular or Nonsingular Variance-Covariance Matrix}.
\newblock \bibinfo{journal}{\emph{Journal of Financial and Quantitative
  Analysis}} \bibinfo{volume}{12}, \bibinfo{number}{3} (\bibinfo{year}{1977}),
  \bibinfo{pages}{347–361}.
\newblock


\bibitem[\protect\citeauthoryear{Frey and Dueck}{Frey and Dueck}{2007}]%
        {aff_prop}
\bibfield{author}{\bibinfo{person}{Brendan~J. Frey} {and}
  \bibinfo{person}{Delbert Dueck}.} \bibinfo{year}{2007}\natexlab{}.
\newblock \showarticletitle{Clustering by Passing Messages Between Data
  Points}.
\newblock \bibinfo{journal}{\emph{Science}} \bibinfo{volume}{315},
  \bibinfo{number}{5814} (\bibinfo{year}{2007}), \bibinfo{pages}{972--976}.
\newblock


\bibitem[\protect\citeauthoryear{Gu, Kelly, and Xiu}{Gu et~al\mbox{.}}{2019}]%
        {gu-2019-autoen-asset}
\bibfield{author}{\bibinfo{person}{Shihao Gu}, \bibinfo{person}{Bryan~T.
  Kelly}, {and} \bibinfo{person}{Dacheng Xiu}.}
  \bibinfo{year}{2019}\natexlab{}.
\newblock \showarticletitle{Autoencoder Asset Pricing Models}.
\newblock \bibinfo{journal}{\emph{SSRN Electronic Journal}}
  (\bibinfo{year}{2019}).
\newblock
\urldef\tempurl%
\url{https://doi.org/10.2139/ssrn.3335536}
\showDOI{\tempurl}


\bibitem[\protect\citeauthoryear{Heaton, Polson, and Witte}{Heaton
  et~al\mbox{.}}{2016}]%
        {heaton-2016-deep-learn-finan}
\bibfield{author}{\bibinfo{person}{J.~B. Heaton}, \bibinfo{person}{N.~G.
  Polson}, {and} \bibinfo{person}{J.~H. Witte}.}
  \bibinfo{year}{2016}\natexlab{}.
\newblock \showarticletitle{Deep Learning for Finance: Deep Portfolios}.
\newblock \bibinfo{journal}{\emph{Applied Stochastic Models in Business and
  Industry}} \bibinfo{volume}{33}, \bibinfo{number}{1} (\bibinfo{year}{2016}),
  \bibinfo{pages}{3--12}.
\newblock
\urldef\tempurl%
\url{https://doi.org/10.1002/asmb.2209}
\showDOI{\tempurl}


\bibitem[\protect\citeauthoryear{Isogai}{Isogai}{2014}]%
        {isogai2014}
\bibfield{author}{\bibinfo{person}{Takashi Isogai}.}
  \bibinfo{year}{2014}\natexlab{}.
\newblock \showarticletitle{{Clustering of Japanese stock returns by recursive
  modularity optimization for efficient portfolio diversification*}}.
\newblock \bibinfo{journal}{\emph{Journal of Complex Networks}}
  \bibinfo{volume}{2}, \bibinfo{number}{4} (\bibinfo{date}{07}
  \bibinfo{year}{2014}), \bibinfo{pages}{557--584}.
\newblock


\bibitem[\protect\citeauthoryear{Isogai}{Isogai}{2016}]%
        {isogai2016}
\bibfield{author}{\bibinfo{person}{Takashi Isogai}.}
  \bibinfo{year}{2016}\natexlab{}.
\newblock \showarticletitle{Building a dynamic correlation network for
  fat-tailed financial asset returns}.
\newblock \bibinfo{journal}{\emph{Applied Network Science}}
  \bibinfo{volume}{1}, \bibinfo{number}{7} (\bibinfo{date}{08}
  \bibinfo{year}{2016}).
\newblock


\bibitem[\protect\citeauthoryear{Isogai}{Isogai}{2017}]%
        {isogai2017}
\bibfield{author}{\bibinfo{person}{Takashi Isogai}.}
  \bibinfo{year}{2017}\natexlab{}.
\newblock \showarticletitle{Dynamic correlation network analysis of financial
  asset returns with network clustering}.
\newblock \bibinfo{journal}{\emph{Applied Network Science}}
  \bibinfo{volume}{2}, \bibinfo{number}{1} (\bibinfo{date}{05}
  \bibinfo{year}{2017}).
\newblock
Issue 8.


\bibitem[\protect\citeauthoryear{Kingma and Welling}{Kingma and
  Welling}{2013}]%
        {kingma-2013-auto-encod}
\bibfield{author}{\bibinfo{person}{Diederik~P Kingma} {and}
  \bibinfo{person}{Max Welling}.} \bibinfo{year}{2013}\natexlab{}.
\newblock \showarticletitle{Auto-Encoding Variational Bayes}.
\newblock \bibinfo{journal}{\emph{CoRR}} (\bibinfo{year}{2013}).
\newblock
\showeprint[arxiv]{1312.6114}~[stat.ML]
\urldef\tempurl%
\url{http://arxiv.org/abs/1312.6114v10}
\showURL{%
\tempurl}


\bibitem[\protect\citeauthoryear{Ledoit and Wolf}{Ledoit and Wolf}{2003}]%
        {varcovPort}
\bibfield{author}{\bibinfo{person}{Olivier Ledoit} {and}
  \bibinfo{person}{Michael Wolf}.} \bibinfo{year}{2003}\natexlab{}.
\newblock \showarticletitle{Improved estimation of the covariance matrix of
  stock returns with an application to portfolio selection}.
\newblock \bibinfo{journal}{\emph{Journal of Empirical Finance}}
  \bibinfo{volume}{10}, \bibinfo{number}{5} (\bibinfo{year}{2003}),
  \bibinfo{pages}{603 -- 621}.
\newblock


\bibitem[\protect\citeauthoryear{Liu, Mulvey, and Zhao}{Liu
  et~al\mbox{.}}{2015}]%
        {liu-2015-semip-graph}
\bibfield{author}{\bibinfo{person}{Han Liu}, \bibinfo{person}{John Mulvey},
  {and} \bibinfo{person}{Tianqi Zhao}.} \bibinfo{year}{2015}\natexlab{}.
\newblock \showarticletitle{A Semiparametric Graphical Modelling Approach for
  Large-Scale Equity Selection}.
\newblock \bibinfo{journal}{\emph{Quantitative Finance}} \bibinfo{volume}{16},
  \bibinfo{number}{7} (\bibinfo{year}{2015}), \bibinfo{pages}{1053--1067}.
\newblock
\urldef\tempurl%
\url{https://doi.org/10.1080/14697688.2015.1101149}
\showDOI{\tempurl}


\bibitem[\protect\citeauthoryear{Polson and Tew}{Polson and Tew}{2000}]%
        {baysianPort}
\bibfield{author}{\bibinfo{person}{Nicholas~G. Polson} {and}
  \bibinfo{person}{Bernard~V. Tew}.} \bibinfo{year}{2000}\natexlab{}.
\newblock \showarticletitle{Bayesian Portfolio Selection: An Empirical Analysis
  of the S\&P 500 Index 1970-1996}.
\newblock \bibinfo{journal}{\emph{Journal of Business \& Economic Statistics}}
  \bibinfo{volume}{18}, \bibinfo{number}{2} (\bibinfo{year}{2000}),
  \bibinfo{pages}{164--173}.
\newblock


\bibitem[\protect\citeauthoryear{Talih and Hengartner}{Talih and
  Hengartner}{2005}]%
        {talih-2005-struc-learn}
\bibfield{author}{\bibinfo{person}{Makram Talih} {and} \bibinfo{person}{Nicolas
  Hengartner}.} \bibinfo{year}{2005}\natexlab{}.
\newblock \showarticletitle{Structural Learning With Time-Varying Components:
  Tracking the Cross-Section of Financial Time Series}.
\newblock \bibinfo{journal}{\emph{Journal of the Royal Statistical Society:
  Series B (Statistical Methodology)}} \bibinfo{volume}{67},
  \bibinfo{number}{3} (\bibinfo{year}{2005}), \bibinfo{pages}{321--341}.
\newblock
\urldef\tempurl%
\url{https://doi.org/10.1111/j.1467-9868.2005.00504.x}
\showDOI{\tempurl}


\end{thebibliography}

\end{document}